\documentclass[twoside,11pt]{article}

\usepackage{jmlr2e}
\usepackage{amsmath}
\usepackage{amssymb}
\usepackage{bbold}
\usepackage{graphicx}
\usepackage{tikz}
\usepackage{subcaption}

\jmlrheading{0}{0000}{0-00}{12/18}{00/00}{00000}{Jakob Knollm\"uller and Torsten A. En\ss lin}

\ShortHeadings{Encoding prior knowledge in the structure of the likelihood }{Knollm\"uller and En\ss lin}
\firstpageno{1}

\begin{document}

\title{Encoding prior knowledge in the structure of the likelihood}

\author{\name Jakob  Knollm\"uller \email jakob@mpa-garching.mpg.de
		\AND
		\name Torsten A. En\ss lin \email ensslin@mpa-garching.mpg.de\\
       \addr Max-Planck-Institut f\"ur Astrophysik, Karl-Schwarzschildstr.~1 \\
       85748 Garching, Germany\\
       Ludwig-Maximilians-Universit\"at M\"unchen, Geschwister-Scholl-Platz{\small{}~}1\\
		80539 Munich, Germany}

\editor{ }
\maketitle

\begin{abstract}
The inference of deep hierarchical models is problematic due to strong dependencies between the hierarchies. We investigate a specific transformation of the model parameters based on the multivariate distributional transform. This transformation is a special form of the reparametrization trick, flattens the hierarchy and leads to a standard Gaussian prior on all resulting parameters. The transformation also transfers all the prior information into the structure of the likelihood, hereby decoupling the transformed parameters a priori from each other. A variational Gaussian approximation in this standardized space will be excellent in situations of relatively uninformative data. Additionally, the curvature of the log-posterior is well-conditioned in directions that are weakly constrained by the data, allowing for fast inference in such a scenario. In an example we perform the transformation explicitly for Gaussian process regression with a priori unknown correlation structure. Deep models are inferred rapidly in highly and slowly in poorly informed situations. The flat model show exactly the opposite performance pattern. A synthesis of both, the deep and the flat perspective, provides their combined advantages and overcomes the individual limitations, leading to a faster inference.
\end{abstract}


\section{Introduction}
Hierarchical Bayesian models make it possible to express the complex relations in real systems by combining a priori domain knowledge with data. The prior knowledge is updated by the observed data to obtain information about the system at hand. Such models can exhibit deep hierarchies, relating a large number of conceptually distinct parameters in non-trivial fashions.
 The inference of the posterior parameters can be extremely problematic, especially for large and complex models due to strong dependencies between the quantities and the resulting numerical stiffness.
  A way to overcome such limitations is to perform coordinate transformations of the parameters to less interdependent ones. In the context of Hamiltonian Monte Carlo (HMC) techniques it was proposed to perform a linear coordinate transformation \citep{HMCWhitening} to decouple the parameters, which in the discussed case leads to a white, standard Gaussian prior for the new parameters. This way the numerical performance was increased. Another, more general transformation scheme was proposed by \citet{HMCHierarchy} to flatten the deep hierarchical structure, to decouple the parameters by introducing auxiliary parameters or performing a reparametrization of existing ones.
  The same kind of transformation is also known as reparametrization trick \citep{AEVB} to learn the parameters of an approximate distribution. It is used in Automatic Differentiation Variational Inference \citep{ADVI} to transform the original model into a standardized space, where a variational approximation is conducted.
  In this paper we  discuss this transformation for general hierarchical Bayesian models and numerical implications of performing variational approximations in these transformed parameters in terms of fidelity and convergence.
  
  We derive this standardizing transformation in two steps. First we transform the original parameters of the deep hierarchy model to independent, uniformly distributed parameters, using the multivariate distributional transform \citep{DistributionalTransform}. 
  This step already removes the deep hierarchy and introduces a uniform prior. The uniform prior is problematic for many inference schemes as it limits the parameter space to the unit interval but does not provide further gradient information for the parameters. Thus, in a second step we then transform the uniform parameters to a Gaussian parameters with unit variance and vanishing mean. The overall transformation is then a non-linear, deterministic machinery which relates uniform, white Gaussian parameters to the original parameters of the deep model. The prior information of the deep model is stored in the structure of the transformation itself. The Gaussian prior allows us to make quantitative statements about the conditioning of the curvature of the log-posterior in different scenarios, which largely determines the difficulty numerical  inference schemes face.  The transformation leads to an optimal conditioning in parameter directions that are only poorly constrained by the data. Directions that are highly constrained by the data result in bad conditioning in this transformed space. Although the transformation discussed does not change the statistical model, inference schemes that involve approximations do depend on the choice of the coordinate system.

As an illustrating numerical example we discuss a Gaussian process regression with also unknown correlation structure. To infer the correlation structure as well statistical homogeneity and isotropy is assumed. The correlation structure can then be expressed in terms of a one-dimensional power spectrum that fully specifies it and has to be inferred together with the signal. To be specific about the statistical process generating the signal we assume the signal to be the result of a zero mean Gaussian process with a kernel as implied by the unknown power spectrum. The log-power spectrum is assumed to be generated by a Gaussian process as well, this time with a known smoothness enforcing kernel.
In this scenario we can conduct the identical approximation in both coordinate systems, allowing us to investigate the effect of the transformation on the numerics. We compare the convergence behavior of this deep and highly coupled model with the transformed, flat model in situations with different amounts of data. Here we find that the deep model performs well in highly informed cases and struggles in low information scenarios. The flat model behaves the other way round. Alternating between the two perspectives in a numerical scheme  overcomes  their individual limitations and provides an overall increased performance.

\subsection{Related works}

\paragraph{Inverse transform sampling}
 \citep{InverseTransform} is used to generate random variables according to an arbitrary distribution from uniform samples. This is done via the inverse  of the cumulative density function (CDF), or quantile function. Its multivariate generalization is the multivariate distributional transform \citep{DistributionalTransform}, which encodes all the internal  hierarchical dependencies of the model. We will use it to reparametrize the original model to obtain the flat formulation.

\paragraph{The reparametrization trick} allows to infer parameters of complex approximate posterior distributions via variational inference. The problem is to take gradients with respect to the parameters of the approximation as the Kullback-Leibler divergence is only estimated statistically  via samples from the approximation \citep{StochasticVB}. Reparametrizing the coordinates of the problem in a certain way allows to separate the randomness of drawing samples from a deterministic modification by the parameters. This can, for example, be achieved analogously to the inverse transform sampling by using the inverse CDF. The advantage of the reparametrization of the distribution is, that now gradients can be calculated with respect to the parameters of the distribution, which only appear in the deterministic part. It is introduced in the Auto Encoding Variational Bayes (AEVB) algorithm \citep{AEVB} to make the parameters of the approximate distribution part of the network architecture. Samples to approximate the variational bound can then be drawn from some simple distribution, allowing the inference of all parameters.
\paragraph{Automatic Differentiation Variational Inference}(ADVI) performs a transformation of the problem to a set of standardized, real-space parameters where then a variational approximation is conducted \citep{ADVI}. We construct this transformation in terms of the multivariate distributional transform and investigate theoretical and numerical properties of performing approximations in this transformed space.
\paragraph{Normalizing flows}
 are used for non-parametric density estimation \citep{NormalizingFlows,NormalizingFlows1}. To apply those one tries to find a set of transformations of the parameters of a simple distribution, such that the transformed distribution matches the target distribution as closely as possible. Here it is important to keep track of the functional determinants introduced by the transformation, in order to keep the resulting distribution normalized. The learned transformation stores all the complexity of the approximate posterior in a deterministic way, whereas the randomness originates from a simple distribution. Similarly, the transformation captures the complex structure of the deep model and relates it back to a simple prior distribution. Instead of learning a suitable transformation, the standardizing transformation makes use of the structure of the hierarchical model. A method that makes use of both, the reparametrization trick, as well as normalizing flow is the variational inference with normalizing flows \citep{NormalizingFlowsAE}.


\section{Basics and notation}
\subsection{Bayesian inference}
In Bayesian inference the prior knowledge $\mathcal{P}(\theta)$, expressed as a probability density function (PDF) of some quantity $\theta$ should be updated after we obtained data $d$. This data is related to the quantity of interest by a likelihood $\mathcal{P}(d\vert \theta)$ of observing the data, given $\theta$. The updated knowledge is the posterior density $\mathcal{P}(\theta \vert d)$ and is calculated according to Bayes theorem:
\begin{align}
\mathcal{P}(\theta \vert d) = \frac{\mathcal{P}(d\vert \theta) \mathcal{P}(\theta)}{\mathcal{P}(d)}
\end{align}
In order to do so, one has to normalize the joint density $\mathcal{P}(d,\theta) = \mathcal{P}(d\vert \theta)\mathcal{P}(\theta)$ with respect to $\theta$, which is done by division with the evidence $\mathcal{P}(d)$.

An alternative way how to describe probabilities is in terms of their information. Information is the negative logarithm of the distribution, $\mathcal{H}(.) = -\mathrm{ln}\: \mathcal{P}(.) $. Compared to the distributions, which are multiplicative, information is additive. This makes it a more convenient quantity to deal with and they will be adapted later in this work. Bayes theorem in terms of information reads:
\begin{align}
\mathcal{H}(\theta \vert d) &= -\mathrm{ln}(\mathcal{P}(\theta \vert d))\\
&= \mathcal{H}(d\vert \theta) + \mathcal{H}(\theta) - \mathcal{H}(d)
\end{align}
The task of Bayesian inference usually comes down to dealing with the normalization term $\mathcal{H}(d)$. In many cases it is not accessible analytically and sampling techniques or approximate inference has to be applied that avoid the calculation of this term. Those only require all terms depending explicitly on the parameters. In order to shorten the notation we will absorb all terms of constant information, therefore parameter independent, into one single information term $\mathcal{H}_0$ and introduce the symbol $\widehat{=}$ that indicates equality up to parameter independent, constant terms.
\begin{align}
\mathcal{H}(\theta \vert d) =  & \:\mathcal{H}_0  + \mathcal{H}(d\vert \theta) + \mathcal{H}(\theta) \\
\widehat{=} & \: \mathcal{H}(d\vert \theta) + \mathcal{H}(\theta)
\end{align}

\subsection{Variational Inference}
Variational inference \citep{VariationalReview} is a powerful tool to approximate an intractable posterior $ \mathcal{P}(\theta \vert d)$ distribution with simpler distribution $\widetilde{\mathcal{P}}(\theta \vert \varphi)$ parametrized by a set of parameters $\varphi$. This is done by minimizing the variational Kullback-Leibler divergence (KL) \citep{KLdivergence} that is defined as:
\begin{align}
\mathcal{D}_{\mathrm{KL}}\left(\widetilde{\mathcal{P}}(\theta \vert \varphi)\vert \vert \mathcal{P}(\theta \vert d)\right) \equiv &\: \int D\theta \: \widetilde{\mathcal{P}}(\theta\vert\varphi ) \:\mathrm{ln}\left[ \frac{\mathcal{P}(\theta\vert d)}{\widetilde{\mathcal{P}}(\theta\vert\varphi)}\right] \\
\widehat{=} & \: \langle \mathcal{H}(d, \theta) \rangle_{\widetilde{\mathcal{P}}(\theta\vert\varphi) } - \langle \widehat{\mathcal{H}}(\theta\vert\varphi)\rangle_{\widetilde{\mathcal{P}}(\theta\vert\varphi) }
\end{align}
In the second line parameter independent terms are dropped, as they are irrelevant for the minimization. This includes the normalization constant of the posterior, which typically is the origin of the intractability of the posterior distribution. This term is also equivalent to the negative evidence lower bound (ELBO) \citep{Bishop}. 
In order to perform the optimization, we do not have to be able to compute the expectation value explicitly, it is sufficient to be capable of drawing samples from the approximate distribution \citep{StochasticVB}.

We will focus on two kinds of variational approximations, point-like, as well as Gaussian approximations.
It can be argued whether fitting delta distributions $\widetilde{\mathcal{P}}(\theta \vert \varphi) = \delta(\theta-\theta^*)$ are truly a variational method, however minimizing the variational KL and maximizing the posterior distribution results in the same procedure. A point estimate on certain parameters might be justified, especially if they are strongly constrained by the problem and they can be inferred significantly faster compared to more sophisticated approaches.

In cases the uncertainty should not be neglected, a variational Gaussian approximation is a powerful tool to infer posterior quantities, as well as their correlations and uncertainties. Gaussian distributions make it simple to sample from them, and derivatives with respect to their parameters can also be calculated easily, as the following expressions hold \citep{GaussianRevisited}:
\begin{align}
\frac{d\mathcal{D}_{\mathrm{KL}}}{d \bar{\theta}} = \left\langle \frac{d \mathcal{H}(d,\theta)}{ d \theta}\right\rangle_{\mathcal{G}(\theta-\bar{\theta},\Theta)} \\
\Theta^{-1} = \left\langle \frac{d^2 \mathcal{H}(d,\theta)}{ d \theta d\theta^\dagger} \right\rangle_{\mathcal{G}(\theta-\bar{\theta},\Theta)}
\end{align}
We will make use of these properties in the example.
\subsection{Transforming probability densities}
Probability densities are differential quantities and to turn them into probabilities they have to be equipped with the differential of their arguments, which are often not stated explicitly, and be integrated over. Keeping track of the differential is relevant for coordinate transformations, as one has to take the differential volume change into account in form of the Jacobian determinant. 
Assume some transformation $\theta'=f(\theta)$ of $\theta$ to a new set of parameters $\theta'$. The probability distribution $\mathcal{P}(\theta)$ then transforms as follows:   
\begin{align}
\theta' &= f(\theta) \\
\mathcal{P}(\theta) d\theta &= \left\vert \frac{d f^{-1}(\theta')}{d\theta'}\right\vert \mathcal{P}(\theta') d\theta' \\
&= \mathcal{P}'(\theta')d\theta'\text{ .}
\end{align}
The vertical lines $\vert \: . \: \vert$ indicate the absolute value of the determinant of the matrix expression inside.
The new distribution $\mathcal{P}'(\theta')$ is the combination of the functional determinant of the transformation and the old distribution of the transformed argument. This way the new distribution is properly normalized. 
\section{Multivariate distributional transform and standard Gaussian priors}
We want to focus on the  transformation that transforms a continuous distribution to a uniform distribution on the unit interval. This is achieved by the quantile function, also known as the inverse cumulative density function (CDF). From the uniformly distributed variables we can then transform to white, standard Gaussian coordinates with the CDF of this Gaussian. In the one dimensional case, the quantile transformation reads:
\begin{align}
\theta_1 &=\mathcal{F}_{\mathcal{P}(\theta_1)}^{-1}(u_1) \:  \text{ ,where }\\
 \mathcal{F}_{\mathcal{P}(\theta_1)}(\theta_1) &\equiv \int_{-\infty}^{\theta_1} d\theta_1' \mathcal{P}(\theta_1') \text{ .}
\end{align}
The first equation is equivalent to inverse transform sampling \citep{InverseTransform} and the second equation defines the CDF. The derivative of the CDF with respect to its argument is therefore again the original PDF.

In general the prior $\mathcal{P}(\theta)$ with $\theta = (\theta_1 \dots ,\theta_n)^\mathrm{T}$ can be expressed in terms of its hierarchical structure
\begin{align}
 \mathcal{P}(\theta) = \mathcal{P}(\theta_n\vert \theta_1\dots \theta_{n-1}) \dots \mathcal{P}(\theta_2\vert \theta_1)\mathcal{P}(\theta_1) \text{.}
 \end{align}
From this separation we can build up iteratively the the transformation to the hierarchical parameters $\theta$ from a set of uniformly distributed parameters $u$.
\begin{align}
\theta_1 = & \:\mathcal{F}_{\mathcal{P}(\theta_1)}^{-1}(u_1) \\
\theta_2 =& \:\mathcal{F}_{\mathcal{P}(\theta_2\vert \theta_1)}^{-1}(u_2) \\
&\vdots \nonumber \\
\theta_n =& \:\mathcal{F}_{\mathcal{P}(\theta_n \vert \theta_1,\theta_2 \dots , \theta_{n-1} )}^{-1}( u_n) 
\end{align}
This is the multivariate distributional transform \citep{DistributionalTransform}
for $u_i$ being drawn from the uniform distribution $\mathcal{U}(u_i)$ within the interval $\left[ 0,1\right]$. From this parametrization we can then change to the white, standard Gaussian distribution in a second step. $\mathcal{G}(\xi, \mathbb{1})$ with uniform, diagonal covariance and vanishing mean. In order to be able to find this transformation in practice, it is necessary that one has explicit access to the hierarchical structure of the model and that CDF's either are available or can be approximated efficiently. This limits its applicability to some extent, but even deep hierarchical models are typically constructed by combining simple distributions and transformations.

We will now perform the above stated coordinate transformation to the uniformly distributed parameters within the joint PDF $\mathcal{P}(d,\theta)$ of the data and parameters for a given likelihood $\mathcal{P}(d\vert \theta)$ and prior distribution $\mathcal{P}(\theta)$ via 
$\mathcal{P}(d,\theta) = \mathcal{P}(d\vert \theta)\mathcal{P}(\theta)$. 

\begin{align}
\mathcal{P}(d\vert \theta) \mathcal{P}(\theta) d\theta =&\: \mathcal{P}(d\vert \theta)\left[ \mathcal{P}(\theta_1) \dots \mathcal{P}(\theta_n\vert \theta_1\dots \theta_{n-1})\right] \left\vert \frac{d\theta}{du}\right\vert du \\
=& \: \mathcal{P}(d\vert \theta) \left[ \frac{d}{d\theta_1} \mathcal{F}_{\mathcal{P}(\theta_1)}(\theta_1)\dots \frac{d}{d\theta_n}\mathcal{F}_{\mathcal{P}(\theta_n\vert\theta_1 \dots \theta_{n-1})}(\theta_n) \right] \left\vert\frac{d\theta}{du}\right\vert du\\
=& \:  \mathcal{P}\left(d\vert\mathcal{F}_{\mathcal{P}(\theta)}^{-1}(u)\right) \bigg[ \frac{d}{d\theta_1} \mathcal{F}_{\mathcal{P}(\theta_1)}\left( \mathcal{F}^{-1}_{\mathcal{P}(\theta_1)}(u_1)\right) \nonumber \\ &\dots
\frac{d}{d\theta_n}\mathcal{F}_{\mathcal{P}(\theta_n\vert\theta_1 \dots \theta_{n-1})}\left( \mathcal{F}^{-1}_{\mathcal{P}(\theta_n\vert\theta_1 \dots \theta_{n-1})}(u_n)\right) \bigg] \left\vert\frac{d\theta}{du}\right\vert du \\
=& \:  \mathcal{P}\left(d\vert\mathcal{F}_{\mathcal{P}(\theta)}^{-1}(u)\right) \left[ \frac{du_1}{d\theta_1} \dots \frac{du_n}{d\theta_n}\right] \left\vert\frac{d\theta}{du}\right\vert du\\
=& \:  \mathcal{P}\left(d\vert\mathcal{F}_{\mathcal{P}(\theta)}^{-1}(u)\right) \left\vert\frac{d u}{d\theta}\right\vert \left\vert\frac{d\theta}{du}\right\vert du\\
=& \:   \mathcal{P}\left(d\vert\mathcal{F}_{\mathcal{P}(\theta)}^{-1}(u)\right) du
\end{align}
Here, we first expanded the prior probability into a hierarchical structure and substituted to an integral over the uniform parameters. We then expressed those individual prior probabilities in terms of derivatives of the CDF. Inserting the transformations for each parameter we obtained identity operations by construction. What remained is the product of the derivatives of the new parameters with respect to the old ones that exactly canceled the Jacobian determinant of the transformation. The uniform prior is implicitly present in the last expression, as the parameters $u$ are only defined on the unit interval, where the uniform distribution takes a value of one. 

For numerical purposes this coordinate space is inconvenient to perform inference due to the compact support of the uniform distribution.  To avoid this, we will transform from these uniform parameters into a set of independent Gaussian parameters of unit variance and zero mean. This coordinate transformation is again done via the CDF. It takes the following form:
\begin{align}
u =& \mathcal{F}_{\mathcal{G}(\xi,\mathbb{1})}(\xi) \\
 =& \: \frac{1}{2} + \frac{1}{2} \mathrm{erf}\left( \frac{\xi}{\sqrt{2}}\right) \label{eq:CDFGaussian} \text{\quad ,}
\end{align}
where the error function is defined as 
\begin{align}
\mathrm{erf}(x) = \frac{2}{\sqrt{\pi}} \int_{0}^x e^{-t^2} dt \text{\quad , }
\end{align}
 and we adopt the convention that scalar functions are applied to vectors component-wise. Performing the transformation explicitly yields
\begin{align}
 \mathcal{P}\left(d\vert\mathcal{F}_{\mathcal{P}(\theta)}^{-1}(u)\right) du= & \: \mathcal{P}\left(d\vert\mathcal{F}_{\mathcal{P}(\theta)}^{-1}(u)\right) \frac{du}{d\xi} d\xi \\
=& \:  \mathcal{P}\left(d\vert\mathcal{F}_{\mathcal{P}(\theta)}^{-1}\circ \mathcal{F}_{\mathcal{G}(\xi,\mathbb{1})}(\xi) \right)\left[ \frac{d}{d\xi} \mathcal{F}_{\mathcal{G}(\xi,\mathbb{1})}(\xi) \right] d\xi\\
=& \:  \mathcal{P}\left(d\vert\mathcal{F}_{\mathcal{P}(\theta)}^{-1}\circ \mathcal{F}_{\mathcal{G}(\xi,\mathbb{1})}(\xi) \right){\mathcal{G}(\xi,\mathbb{1})} d\xi \text{\quad .}
\end{align}
The overall standardizing transformation $ \mathcal{C}_{\mathcal{P}(\theta)}(\xi)$ of the deep hierarchical parameters $\theta$ to the white, standard Gaussian parameters $\xi$ is therefore the composition, indicated by $\circ$, of the two individual transformations. 
\begin{align}
\theta =&\mathcal{F}_{\mathcal{P}(\theta)}^{-1}\circ \mathcal{F}_{\mathcal{G}(\xi,\mathbb{1})}(\xi) \\
\equiv &\: \mathcal{C}_{\mathcal{P}(\theta)}(\xi)
\end{align}
Finally, we can rewrite the original probability in terms of the new parameters.
\begin{align}
\mathcal{P}(d\vert\theta)\mathcal{P}(\theta) d\theta=&  \mathcal{P}\left(d\vert \mathcal{C}_{\mathcal{P}(\theta)}(\xi) \right){\mathcal{G}(\xi,\mathbb{1})} d\xi
\end{align}
Explicit examples for a simple hierarchical model and multivariate Gaussian prior distributions are given in Appendices \ref{app:simple} and \ref{app:multivariate}, reproducing the reparametrization trick.

\section{Approximations of the transformed distributions}
Approximating posterior distributions allows to infer posterior quantities even for large models within reasonable computational effort. In general it matters in which coordinate system the approximation is conducted. We will discuss the impact of this transformation on two popular approximations in the standardized parameters. The first one will be the maximum posterior (MAP) estimate that is obtained by minimizing the information of the posterior with respect to the parameters. The second approach is to perform a variational approximation with a Gaussian distribution in the standardized coordinates.
\subsection{Maximum Posterior}
\label{sec:MAP}
A maximum posterior approximation is cheap to compute and can provide meaningful results, if the parameters are constrained reasonably well and uncertainties are small. Conceptually, the true posterior distribution is approximated by a delta distribution at a location that has to be inferred. Minimizing the KL divergence in this case is identical to minimizing the information $\mathcal{H}(\theta \vert d)$ with respect to the parameters $\theta$. Performing the approximation in the standardized coordinate system will not necessarily maximize the posterior in the original parametrization. We can discuss the two limiting cases of uninformative likelihood and extremely constraining data. 

In the case of an uninformative likelihood, maximizing the posterior will be close to maximizing the white, standard Gaussian information. The result will be a delta distribution peaked close to the the origin. This location splits the probability mass in any direction in half. Transforming this distribution back into the original coordinates this location corresponds to the median of the prior distribution rather than the (or a) mode that would be the result of MAP in the original space. Especially for heavily skewed or multi-modal prior distributions the results differ substantially.

In the limiting case of highly informative data the true posterior distribution will be narrowly peaked around the true parameter value. In this situation the essential features of the posterior can be captured by an approximation with a delta distribution, neglecting uncertainty. The true posterior then also transforms almost like a delta distribution, which will also be narrowly peaked around the transformed maximum. In the highly informed case it therefore does not matter much in which coordinates the approximation is conducted. 

Any situation in between the extreme cases will exhibit characteristics of both. In general, performing a MAP approximation in the standard coordinates will push towards median prior configurations, unless the data tells otherwise. This is different to the approximation in original coordinates, which favors maximum prior configurations in the absence of additional information. Towards more conclusive data it becomes irrelevant in which coordinates the posterior is maximized.

\subsection{Variational Gaussian}
A natural choice to approximate the true posterior in the transformed coordinates is the Gaussian distribution. This is demonstrated for ADVI \citep{ADVI}. The transformed prior distribution is just a standard Gaussian. If the data updates the prior only slightly, the true posterior will still be close to a Gaussian distribution, which is captured well by the approximation. The strength of variational inference is to take the uncertainty of the problem into account and thereby prevents over-fitting to some extent. This is especially important if the uncertainty is high, which is the case for uninformative data. This is exactly the situation where the variational Gaussian approximation in the standardized space captures the true uncertainty of the actual posterior the best. 

A variational Gaussian approximation will also approximate the true posterior well if the likelihood in the transformed coordinates is close to a Gaussian distribution in the parameters, as combining a Gaussian likelihood and prior results in another Gaussian.

For well constrained parameters this approximation will behave similarly to the MAP approximation, as discussed before. If the likelihood introduces multi-modality into the posterior, the Gaussian approximation will choose one mode and approximate the true posterior locally. In these situations one might consider more flexible distributions to approximate the posterior, for example a Gaussian mixture, as demonstrated in \citet{ADVI}.

Overall, the variational Gaussian approximation of the true posterior in the standardized space will be excellent if the data modifies the posterior only slightly. 

\section{Optimization and Conditioning}
In order to perform the approximate inference, we do have to optimize a target functional numerically. This problem might be a non-convex optimization and we are not guaranteed to find a global optimum. Here we will discuss the convergence properties of local optimization procedures of the MAP and variational Gaussian approximation in the standardized coordinates. To perform the optimization we do only have access to local information in terms of derivatives of the loss function at the current position in parameter space. Here the negative gradient shows in the direction of the steepest descent and the curvature informs about how the terrain is changing along the different directions. This information is used in a number of Newton and quasi-Newton algorithms to minimize the target functional. The numerical difficulty is encoded in the condition number of the curvature, the ratio of the absolutes of its largest and smallest eigenvalue. The larger the spread of the eigenvalues, the harder the problem is to solve numerically.
For non-convex problems the curvature might exhibit negative eigenvalues, especially far from a minimum. In this case  a Newton optimization breaks down, as the step will be performed in the wrong direction. A way to overcome this is to approximate the true curvature with a quadratic approximation to the curvature with the Gauss-Newton method that is always positive definite. For our discussion we will consider only convex curvature. Otherwise the same argumentation holds for an approximate Gauss-Newton curvature.

\paragraph{Conditioning of general models:}
The general curvature of the information for a deep hierarchical model reads:
\begin{align}
C = \frac{d^2}{d\theta d\theta^\dagger} \mathcal{H}(d\vert \theta) + \frac{d^2}{d\theta d\theta^\dagger} \mathcal{H}(\theta)
\end{align}
The conditioning of this curvature will entirely depend on the concrete model, but a number of general properties influence the conditioning strongly. Large absolute entries in the curvature matrix will contribute to a bad conditioning, wherever they occur. The second derivative with respect to two parameters will be large if the relevant terms themselves are highly informative, as well as if the interaction of the parameters within these terms is strong. Highly informative and strongly coupled terms will therefore contribute to bad conditioning.  Not only individual, large entries are problematic, but also a large number of relatively small entries associated with one individual parameter. This case is discussed in \citet{HMCHierarchy} and illustrated there in form of a high dimensional funnel.

\paragraph{Preconditioning:}
A common technique to reduce the condition number of a linear problem is preconditioning \citep{AgonizingPain}. The idea is to have an approximation of the original problem that has an easily accessible inverse. This approximation is pulled out of the initial matrix, taking already care of the largest and smallest eigenvalues. It remains to solve a better conditioned problem. For example, we want to have the inverse of a matrix that consists of a simple, invertible, and dominant contribution $B$ plus a small modification $A$. The condition number is therefore dominated by the matrix $B$. A way to precondition this problem is to pull $B$ out of it:
\begin{align}
(A + B)^{-1} = B^{-1}(AB^{-1} + \mathbb{1})^{-1} 
\end{align} 
Now it remains to numerically invert $(AB^{-1} + \mathbb{1})^{-1}$  instead of the initial problem. Here the smallest eigenvalue is bounded by $1$ and the largest eigenvalue relates to the largest eigenvalue of $AB^{-1}$ (plus one), which is smaller than the product of the largest eigenvalues of $A$ and $B^{-1}$. As $A$ is only a small modification, its largest eigenvalue will also be small, and therefore pulling out the dominant contribution $B$ leads to a better conditioned problem. 

\paragraph{Curvature of MAP and Variational Gaussian:}
The curvature of the MAP approximation, which is used in the Laplace approximation to obtain an uncertainty estimate, is the second derivative of the information. It reads in the standardized coordinates:
\begin{align}
C_\mathcal{H} = \frac{d^2}{d \xi d \xi^\dagger} \mathcal{H}(d\vert \xi) + \mathbb{1}
\end{align} 
For a variational Gaussian approximation we can pull derivatives with respect to the mean inside the expectation value and take the derivative with respect to the original parameters \citep{GaussianRevisited}:

\begin{align}
C_{\mathcal{D}_{\mathrm{KL}}} =\left \langle\frac{d^2}{d \xi d \xi^\dagger} \mathcal{H}(d\vert \xi) + \mathbb{1} \right\rangle_{\mathcal{G}(\xi-m_\xi,D_\xi)}
\end{align} 
The curvature for the mean of the Gaussian approximation is the mean of the information curvature over the approximate distribution. This structure allows us to investigate the overall conditioning of the curvature in terms of the largest and smallest eigenvalues introduced by the likelihood. The structure of both curvatures above correspond to the one of the preconditioned problem in the previous paragraph and we can discuss them in the same way.

\paragraph{Conditioning:} 
The above considerations allow us to make statements on the conditioning of the problem, as the condition number is given by
\begin{align}
\kappa = \frac{\lambda_{\mathrm{max}}+1}{\lambda_{\mathrm{min}}+1} \text.
\end{align}
Here  $\lambda_{\mathrm{max}}$  and $\lambda_{\mathrm{min}}$ are the largest and smallest eigenvalues of the likelihood information curvature.
The magnitude of the eigenvalues relates to the uncertainty in the corresponding eigendirections of the posterior. This property is used for the Laplace approximation. Large eigenvalues indicate low posterior variance, and therefore well constrained parameter directions, and vice versa. Certain directions might not be constrained by the data at all and the posterior uncertainty is the prior one, which is indicated by $\lambda_{\mathrm{min}}=0$. Because of this, the smallest eigenvalue of the full curvature cannot become smaller than $1$.

The overall conditioning of the problem therefore mainly depends on $\lambda_{\mathrm{max}}$. The larger it is, the worse the overall conditioning. The inference of the model will therefore be faster for less informative data. If the smallest eigenvalue $\lambda_{\mathrm{min}}$ is significantly above one, the data constrains all parameters well and the prior will not have much influence on the conditioning.

The bad conditioning for highly informative data can be explained by the structure of this likelihood. All parameters are directly involved into explaining the data. Within the likelihood the parameters are to some extent degenerate, as several might explain the same features. The prior breaks this degeneracy, favoring certain parameter configurations above others. If the likelihood is now extremely strong, the influence of the prior almost vanishes. The first goal of the algorithm will be to minimize the likelihood, irrespective of the prior plausibility. To restore this plausibility, the optimization has to also minimize the prior contributions. This is now only possible by following a  trajectory that keeps the likelihood almost constant. This quasi-hard-constraint introduces narrow, high-dimensional valleys into the information function the optimization has to navigate through. 

For well constrained parameters we might prefer a MAP approximation. As previously discussed, the resulting estimate should not depend significantly on the coordinates chosen. A way to circumvent the bad conditioning in the highly informed case might be to perform the optimization in the original parameter space with the deep hierarchy.

\section{Numerical example}
To illustrate the above considerations we present a numerical example. We will explore the inference of a linear Gaussian process $s$ with unknown kernel operator $S$ from incomplete data with additive, Gaussian noise $n$ . This is an important problem, as Gaussian processes \citep{krige1951statistical} are widely used to model continuous functions or auto-correlated quantities \citep{GPML}. They are also well-suited for image reconstruction, for example in an astrophysical context \citep{kitaura2008bayesian,IFT, D3PO,oppermann2015estimating, RESOLVE}. The auto-correlation of the process is the result of the properties of the observed system, which might not be known a priori, so it also has to be learned from the data. Parametric \citep{williams1996gaussian}, as well as non-parametric \citep{wilson2013gaussian} models have been proposed to infer the correlation structure. The latter work assumes a mixture of Gaussian profiles, which, in principle, can represent any viable spectral density for a sufficiently large number of basis functions. Alternatively the spectrum itself can also be described using a  Gaussian process for the logarithmic spectrum, ensuring positive definiteness \citep{ensslinfrommert}.  This log-normal prior on the spectral density can, for example, enforce spectral smoothness \citep{smoothpower}. We will base our discussion on this latter description, but the results should hold for any parametrization of the correlation function.

\subsection{Gaussian process with spectral smoothness}

Gaussian processes are defined over continues spaces, and therefore the involved quantities will be functions and linear operators instead of vectors and matrices.
We will additionally consider the presence of a general, linear response function $R$, which can, for example, select out individual locations where we measure the signal, resulting in our data-points. This operator is necessary to relate the continuous signal $s$ to discrete-valued data.
The data is generated according to 
\begin{align}
d = Rs + n \text{.}
\end{align}
This results in a Gaussian likelihood with information of the form:
\begin{align}
\mathcal{H}(d\vert s) = \frac{1}{2} (d-Rs)^\dagger N^{-1} (d-Rs) +\frac{1}{2} \mathrm{ln}\vert 2\pi N \vert
\end{align}
The information of a Gaussian process prior reads:
\begin{align}
\mathcal{H}(s\vert S) = \frac{1}{2} s^\dagger S^{-1}s + \frac{1}{2} \mathrm{ln}\vert 2\pi S \vert
\end{align}
The kernel $S(x,x')$ should be homogeneous, or stationary, and it therefore only depends on the relative distance of two points $S(x-x')$ that allows us to express the correlation as a diagonal operator in the harmonic basis. The additional assumption of isotropy allows for one dimensional kernel functions that only depend on the relative distance $S(\vert x - x'|)$. The correlation structure is then a diagonal operator in the harmonic basis and fully characterized by its spectral density, according to the Wiener-Khintchin theorem \citep{Wiener, Khintchin}. The isotropy assumption implies that the spectral density only depends on the absolute values of the coordinates in the harmonic space. This can be expressed via an isotropy operator $\mathbb{P}$, that distributes a one dimensional power spectrum into the diagonal of the covariance operator in the harmonic space. The relation between the correlation structure in position space $S$ and a one dimensional power spectrum $p$ therefore reads:
\begin{align}
S = \mathbb{F}^\dagger \widehat{\left(\mathbb{P}p\right)}  \mathbb{F}
\end{align}
The hat over $\widehat{\left(\mathbb{P}p\right)} $ indicates the transformation into a diagonal operator and the harmonic transformation is expressed in terms of $\mathbb{F}$. For flat geometries this is the Fourier transformation. In order to enforce the positive definiteness of the correlation structure, $p$ has to be strictly positive, but its values can vary strongly. In many cases one can assume a smooth power spectrum. A suitable choice of a prior to implement these characteristics is a log-normal Gaussian process prior $\mathcal{LN}(p, T)$. The kernel $T$ implements the degree of desired smoothness. As this contains the hard constraint of positivity, we will instead reparametrize the power spectrum in terms of the logarithmic power spectrum $p = e^\tau$, which transforms the hyper-prior to the Gaussian Process prior $\mathcal{G}(\tau,T)$. The total information is obtained by adding up all likelihood, prior, and hyper-prior terms. Disregarding all parameter independent terms, it is:
\begin{align}
 \mathcal{H}(d,s,\tau)\:\widehat{=}\:&\:\frac{1}{2} (d-Rs)^\dagger N^{-1} (d-Rs) \nonumber \\
 &+ \frac{1}{2} s^\dagger  \mathbb{F}^\dagger \widehat{\left(\mathbb{P}e^{-\tau}\right)}  \mathbb{F}s  + \frac{1}{2}\mathrm{Tr}\left(\mathbb{P}\tau\right) +\frac{1}{2} \tau^\dagger T^{-1} \tau 
\end{align}
We will now apply the standardizing transformation to flatten the hierarchy of the Bayesian model. Due to the  assumed statistical homogeneity of the signal, we have access to the eigenbasis of the prior correlation structure. Here $\mathbb{F}$ is the Fourier transformation, which allows us  to take the square root of the eigenvalues to standardize the $s$ parameter as outlined in Appendix \ref{app:multivariate} .
\begin{align}
s = \mathbb{F}\widehat{\left(\mathbb{P} e^{\frac{1}{2}\tau}\right)}\xi
\end{align}
Performing this substitution introduces the dependency on $\tau$ into the likelihood and removes the prior terms depending on  $\tau$. 

The same procedure can be applied to standardize $\tau$ as well, which is also described by a Gaussian process. In order to do so we have to express the smoothness kernel $T$ in terms of its eigenbasis and eigenvalues. Smoothness should be a lack of curvature of $\tau$ on a logarithmic scale. We can therefore describe the inverse kernel as $T^{-1} = \frac{1}{\sigma^2} \Delta ^\dagger \Delta$. The $\Delta$ operator implements the Laplace operator on a logarithmic coordinate system and $\sigma$ the expected deviations from a smooth spectrum. The larger it is, the more curvature is accepted and vice versa. The Laplace operator is diagonal in the associated harmonic domain and the diagonal elements contained the squared harmonic coordinate $l$ of this logarithmic space. 
\begin{align}
\Delta = \mathbb{V}^\dagger l^2 \mathbb{V}
\end{align}
This $\mathbb{V}$ operator is the harmonic transformation in the space of the one dimensional logarithmic power spectrum in logarithmic coordinates.
We can now express $\tau$ in terms of the standard parameters $\zeta$:
\begin{align}
\tau = \mathbb{V} \frac{\sigma}{l^2} \zeta
\end{align}
With both transformation in place, the transformed information of the full problem reads:
\begin{align}
\mathcal{H}(d,\xi,\zeta) \:\widehat{=} \: &\:\frac{1}{2} \left(d-R \mathbb{F}\widehat{\left(\mathbb{P}e^{\frac{1}{2}\mathbb{V}\frac{\sigma}{l^2} \zeta}\right)}\xi\right)^\dagger N^{-1}\left(d-R \mathbb{F}\widehat{\left(\mathbb{P}e^{\frac{1}{2}\mathbb{V}\frac{\sigma}{l^2} \zeta}\right)}\xi\right) \nonumber \\
&+\frac{1}{2} \xi^\dagger \mathbb{1} \xi  +\frac{1}{2} \zeta^\dagger \mathbb{1} \zeta 
\end{align}
Now the entire prior knowledge, such as the concepts of homogeneity, isotropy,  spectral and spatial smoothness, and positivity, are absorbed into the likelihood. This now implements the problem of the inference of a Gaussian process with unknown correlation structure in form of a characteristic and generative sequence of linear and nonlinear operations between a priori white parameters. The steps to perform the transformation were technical, but straight forward.
\subsubsection{Inference}
So far we only formulated the identical problem in two equivalent ways, a deep hierarchical model and a flat hierarchical model with a  standard Gaussian white prior, but we still have to perform the inference.
In this case the performed coordinate transformations were fully linear. This way we can perform the same approximation in both coordinates systems and the resulting distributions will be transformed versions of each other. 
We will minimize the variational KL divergence between the true posterior distribution and an approximate distribution. The approximation will be a product of a Gaussian for the signal and a point-estimate for the power spectrum. This illustrates a variational Gaussian approximation, as well as point estimates. 

The approximate distribution reads:
\begin{align}
\mathcal{P}(s,\tau \vert m,D,\tau^*) = \widetilde{\mathcal{P}}(s,\tau) \equiv\mathcal{G}(s-m,D) \delta(\tau-\tau^*)
\end{align}
The task is to adjust the parameters $m$, $D$ and $\tau^*$ such that the KL divergence between this distribution and the true posterior is minimized. Up to parameter independent, constant terms the KL divergence reads:
\begin{align}
\mathcal{D}_{\mathrm{KL}}\left(\mathcal{P}(s,\tau\vert d)\vert \vert\widetilde{\mathcal{P}}(s,\tau) \right)\widehat{=}& \langle \mathcal{H}(s,\tau\vert d)\rangle_{\widetilde{\mathcal{P}}(s,\tau) } - \langle \widetilde{\mathcal{H}}(s,\tau) \rangle_{\widetilde{\mathcal{P}}(s,\tau) }\\
\widehat{=} &\: \frac{1}{2} \left\langle  (d-Rs)^\dagger N^{-1} (d-Rs)+ \frac{1}{2} s^\dagger  \mathbb{F}^\dagger \widehat{\left(\mathbb{P}e^{-\tau^*}\right)}  \mathbb{F}s  \right\rangle_{\mathcal{G}(s-m,D)} \nonumber \\&+ \frac{1}{2} \mathrm{Tr}\left(\mathbb{P}\tau^*\right) + \frac{1}{2} \tau^{*\dagger} T^{-1} \tau^* -\mathrm{Tr} \:\mathrm{ln} \left[2\pi e D\right]
\end{align}
The last term corresponds to the entropy of the Gaussian distribution and the delta distribution is already integrated out.
Because the information is fully quadratic in $s$, we can directly solve for the posterior covariance, as well as the mean for a given $\tau^*$ \citep{GaussianRevisited}:
\begin{align}
D &= \left( R^\dagger N^{-1} R +\mathbb{F}^\dagger \widehat{\left(\mathbb{P}e^{-\tau^*}\right)}  \mathbb{F}\right)^{-1}\\
j &= R^\dagger N^{-1} d\\
m &= Dj \text{ .}
\end{align}
$\mathcal{P}(s\vert d, \tau^*)= \mathcal{G}(s-m,D)$  is the Wiener filter solution for a given correlation structure $\tau^*$ and it minimizes the KL divergence without the need of a dedicated optimization. The minimization with respect to $\tau^*$  requires the evaluation of the KL divergence and its gradient with respect to these parameters. In order to estimate the Gaussian expectation value we will draw samples from the Gaussian distribution for the current value of $\tau^*$ and perform the optimization on a stochastic estimate of the KL divergence \citep{StochasticVB}.

The inference procedure now iterates between estimating $m$ and $D$ for a given $\tau^*$ and minimizing a stochastic estimate of the KL-divergence with respect to $\tau^*$  for the current parameters of $m$ and $D$.

Analogous terms can be calculated for the standardized coordinates and the inference procedure will be identical.

\subsubsection{Implementation and results}
\paragraph{Setup:}
In the concrete example we will consider a two dimensional Gaussian process drawn from the prior distribution. We generate data by measuring the process at random locations and we additionally add white Gaussian noise. We will present three times the identical setup, varying only the amount of data points. We choose a resolution of $128 \times 128$ pixel. The first scenario will consist of a measurements of all locations, in the second scenario we randomly sample $10\%$, or roughly $1600$ locations, and in the last case we only take $0.5\%$, or $83$ data points. 

The problem is implemented in python using the NIFTy package \citep{Nifty,Nifty3}.
We start the optimization of both models with equivalent initial states. All hyper-parameters of the involved minimizers are the same as well.
 
Using the true underlying kernel, we can compute the posterior mean $m_{\mathrm{wf}}$ of this simpler problem by evaluating the Wiener filter for the correct spectrum. This better informed estimator will serve as our reference point and we compare both methods in terms of the root mean squared errors (RMS), which are defined as:

\begin{align}
\epsilon =\sqrt{(m - m_{\mathrm{wf}})^\dagger(m-m_{\mathrm{wf}})}
\end{align}
 By monitoring this quantity we can discuss the convergence behavior of the different methods in the different scenarios.

We do not track the KL divergence as measure for convergence, as it would require the calculation of the entropy term $\mathrm{Tr} \: \mathrm{ln}[2\pi e D]$. The posterior covariance $D$ has $128^2 \times 128^2$ entries and the eigenbasis is not explicitly available. Evaluating the matrix logarithm scales with the third order in the dimension, requiring $128^6$ computations in every step, making this term numerically not accessible. In order to perform the inference itself it is sufficient to have the operator implicitly available.

The overall convergence of the algorithm is determined by the convergence of the power spectrum parameters, as we immediately have the mean and variance of the Gaussian given this spectrum. In order to discuss the convergence behavior it is important to identify which parts of the power spectrum are prior dominated and which are strongly constrained by the likelihood. In order to do so we have to explore how the data contains information on the correlations of different scales. Our knowledge on the true process realization is limited in two ways, namely  white Gaussian noise and incomplete coverage. 

This noise affects all scales equally, and the relevant quantity to look at is the signal-to-noise ratio. Scales with high power will not be affected much, and are therefore well constrained by the data. If the signal power drops  significantly below the noise power, the prior information will be the dominant term.

The random sampling behaves differently. If we randomly select a small number of positions to measure, their average distance is large and we are mainly informed about the largest scales. We are informed about the small scales by having data points close to each other. The more data points we obtain, the more likely it becomes for them to lie close to another one, providing information on the small scale fluctuations.

In our case both effects will be superimposed, but the sparse sampling of data will be the main contribution. With it we will stir between prior and likelihood dominance.

\begin{figure}
	\centering
	\begin{subfigure}[b]{\textwidth}
		\vspace{-0.5cm}
		\hspace*{1.1cm}                                                           
		\includegraphics[width=\textwidth]{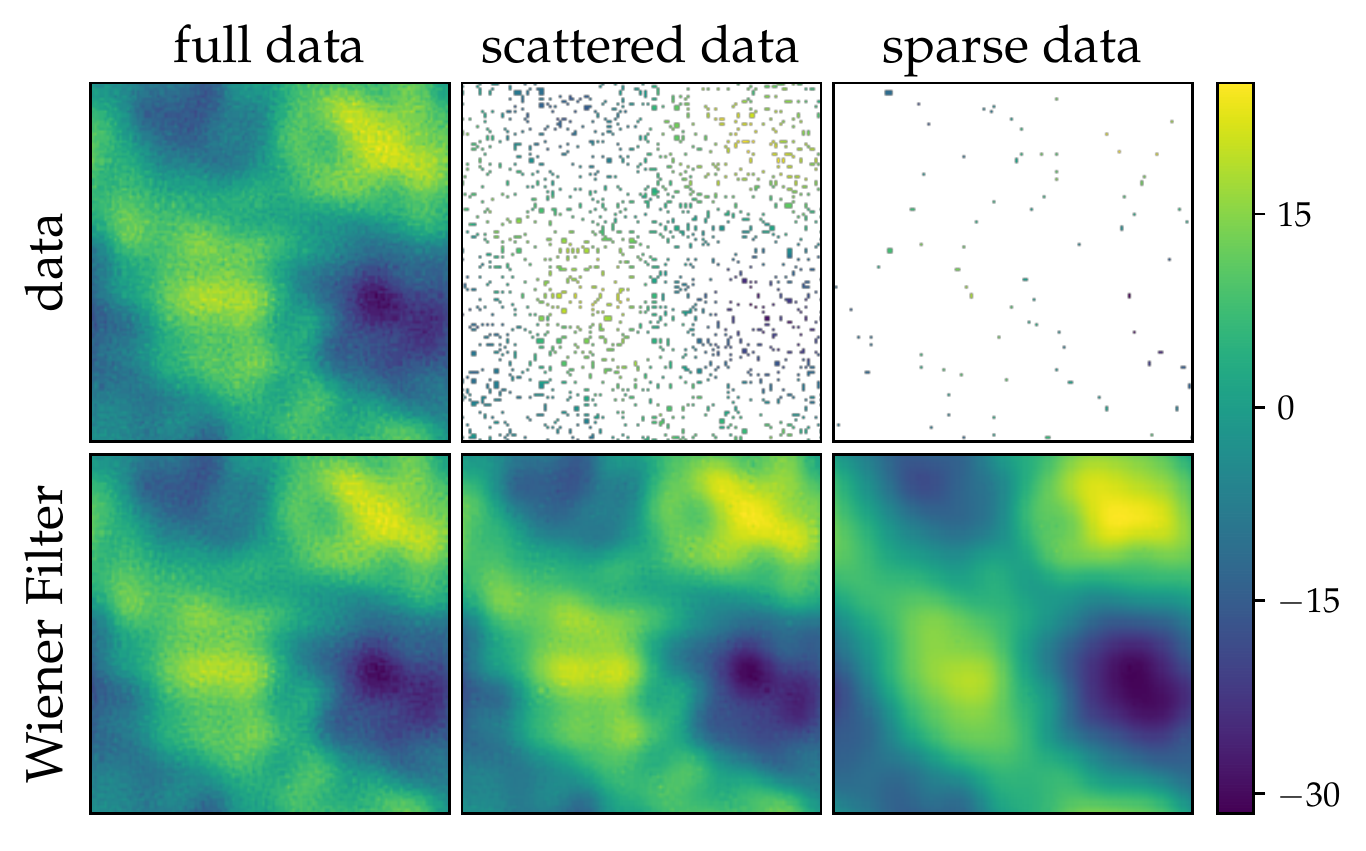}
	\end{subfigure}\\
	\begin{subfigure}[b]{\textwidth}
		 \vspace{-0.5cm}
		\includegraphics[width=\textwidth]{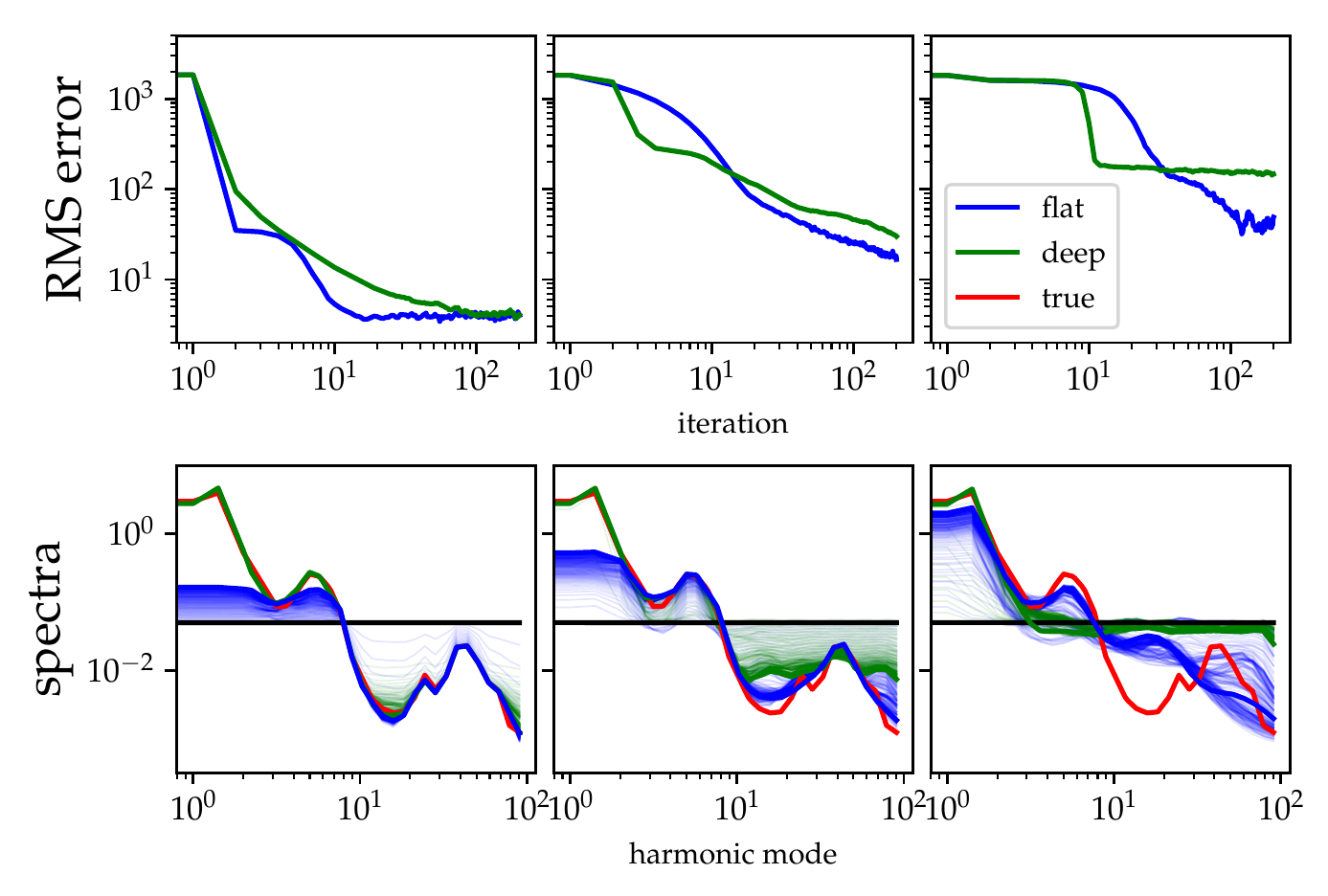}
	\end{subfigure}
	\caption{The setup and results in the three cases. The column corresponds to the data set with full coverage, the second one to the scattered data set, and the last one to the sparse data set. The first row shows the data in these cases, the second row depicts the Wiener filter reconstruction given the correct correlation structure. The third row illustrates the RMS error of the current reconstruction to the Wiener Filter solution for both coordinates and the last row shows the progression of the power spectra during the optimization. }
	\label{fig:critical_filter}
\end{figure}

The results can be seen in Fig. \ref{fig:critical_filter}. The three columns correspond to the three different cases. The first row shows the actual data, the second row the posterior mean for the correct kernel. The third row illustrates the RMS error of the current estimate after each algorithmic update during the minimization. The last row shows the progression of the spectra during the inference.

\paragraph{Observations:} 
\begin{enumerate}
\item The flat hierarchy model converges faster in all cases in terms of RMS error.
\item In recovering the true spectrum, both models have complementary strengths and weaknesses. The deep hierarchy is superior in the data dominated case and the flat hierarchy in the prior dominated case.
\item Alternating  both methods improves convergence with respect to each and the recovery of the true spectrum.
\end{enumerate}

\paragraph{Convergence:} 
The convergence results can be seen in the third row in Fig \ref{fig:critical_filter}. The flat hierarchy model converges faster in all three cases in terms of the RMS error. In the full data case both methods converge to the identical error, but the flat hierarchy requires one order of magnitude less iterations. In the scattered data cases, the deep model is faster at the beginning, but after a certain amount of steps, the flat model outpaces the other one significantly. In the sparse data case, the deep hierarchy seemingly stops converging after an initial drop off, whereas the flat model reaches a significantly lower error. This illustrates the advantages of the approximation in the standardized space for relatively uninformative data. In the scattered data case, the convergence behavior is similar, with slight advantage for the flat model. The problem in this case is that the large scales are well constrained by the data, whereas the small scales should still be prior dominated. Because the deep hierarchy converges fast for well constrained parameters and struggles for the less constrained parameters, the over-all convergence is bottle-necked by the most uninformative parameters. The flat model shows exactly the opposite behavior in the different regimes, and is therefore bottle-necked by the most informed parameters. These behaviors become especially apparent in the convergence of the power spectra.

\paragraph{Spectra:} 
We will now discuss the evolution of the power spectra during the minimization. These are depicted in the last row of Fig. \ref{fig:critical_filter}. Both methods start with spectra at the  same horizontal line. From this location the spectra move into the direction of the true correlation structure. We can track fast movement by a low density of lines and slow movement by a high density. Convergence can be identified by small fluctuations around a common spectrum. We also observe the evolution to slow down and almost stopping at some point. This behavior appears as  a color gradient in the figure, which progressively gets filled stronger. 

The deep hierarchy model is extremely fast in picking up scales, which are well constrained by the data. This can be seen by the density of green lines at the largest scales. In the full data case it immediately jumps close to the true spectrum. For sparser data this is a bit delayed, but still fast. This slowing down cannot only be observed between the cases, but also within the different scales in one setting. The smaller scales are less constraint by the data and for those there are a large number of intermediate lines. In the scattered data case this behavior can be seen very well. There the small scales just slowly drop down from the initial position towards the correct spectrum, bottle-necking the convergence. 

In the sparse data case, this slow down behavior is even more dominant. Everything except the largest scales did not move away significantly from the initial position and the minimization is incapable to provide a result within a reasonable amount of time. 

The overall behavior can be explained by the deep hierarchical structure of the model. We use the current correlation structure to estimate the mean of the process. For parameters the data is good, this current prior correlation structure does not affect its estimate strongly and if data are sparse, the posterior remains close to the prior estimate. This updated process is then used to adjust the correlation structure within the prior distribution. Therefore well constraint scales are immediately set to reasonable values, whereas the weakly constraint modes are already consistent with the current estimate of the power spectrum. The algorithm gets caught in a loop of self-fulfilling prophecies, which cripples down any progress on scales with little data.

The flat hierarchy model behaves to some extent in an opposite manner. It is fast to recover scales which are weakly constraint by the data, but struggles for well informed scales. The more data it is available, the slower the spectrum approaches the true solution. In situations where the deep hierarchy has problems, this flat model recovers the spectra reasonably. Especially in the sparse data case it is capable on capturing the basic features of the true spectrum. Besides the clearly identified secondary bump, it also estimates the slope of the true spectrum correctly. Spectral features on even smaller scales could not be identified due to the lack of information. Here the correct spectrum is smoothly interpolated, as we would expect for a Gaussian process in weakly informed scenarios. The flat model is excellent in combining small amounts of data with prior knowledge. Interestingly, its incapability of recovering the correct spectrum in the high data density case does not reflect in the reduction of the RMS error, where the flat model consistently out-competes its competitor. 

This behavior can be explained by a multiplicative degeneracy within the likelihood. We re-parametrized the original signal in terms of an amplitude and an excitation:
\begin{align}
s = A \xi
\end{align}
This allows to multiply one quantity with some factor, if the other one is divided by it accordingly. In the algorithm we first update the excitation $\xi$  for a given amplitude. In the case of a strong likelihood, the white prior on $\xi$ is almost irrelevant and the excitation will pick up any access power, which is not explained by the initial guess of the amplitude. In the consecutive update of the amplitude itself, this power is hidden inside the excitation and the amplitude cannot pick it up. The likelihood is immediately satisfied and the problem remains in the separation between excitation and amplitude, which is only governed by the priors. The high values in the excitation are incompatible with the prior, which wants to push them down, but it has to act against the extremely strong likelihood. This way, the prior will only slightly push down the values, which releases some power to be picked up by the amplitude. The stronger the likelihood, the slower this process will be and the push from the excitation prior becomes weaker, the progression of the power spectrum will slow down. 

\paragraph{Alternating coordinates:} Both parametrizations have severe limitations, which is best depicted in the spectra for the scattered data.  On large scales, the flat model struggles with internal degeneracy, whereas for small scales the deep model is fighting with self-fulfilling prophecies. Both methods do also have their advantages. The deep model is excellent in the high density data regime, whereas the flat hierarchy allows to recover features even in low information environments.

In this case we can alternate between the coordinates without altering the approximation, due to the linearity of the transformation. In general this is not possible, but as discussed in Sec. \ref{sec:MAP} delta approximations for well constrained parameters can also be optimized in transformed coordinates. 

\begin{figure}
\includegraphics[width=\textwidth]{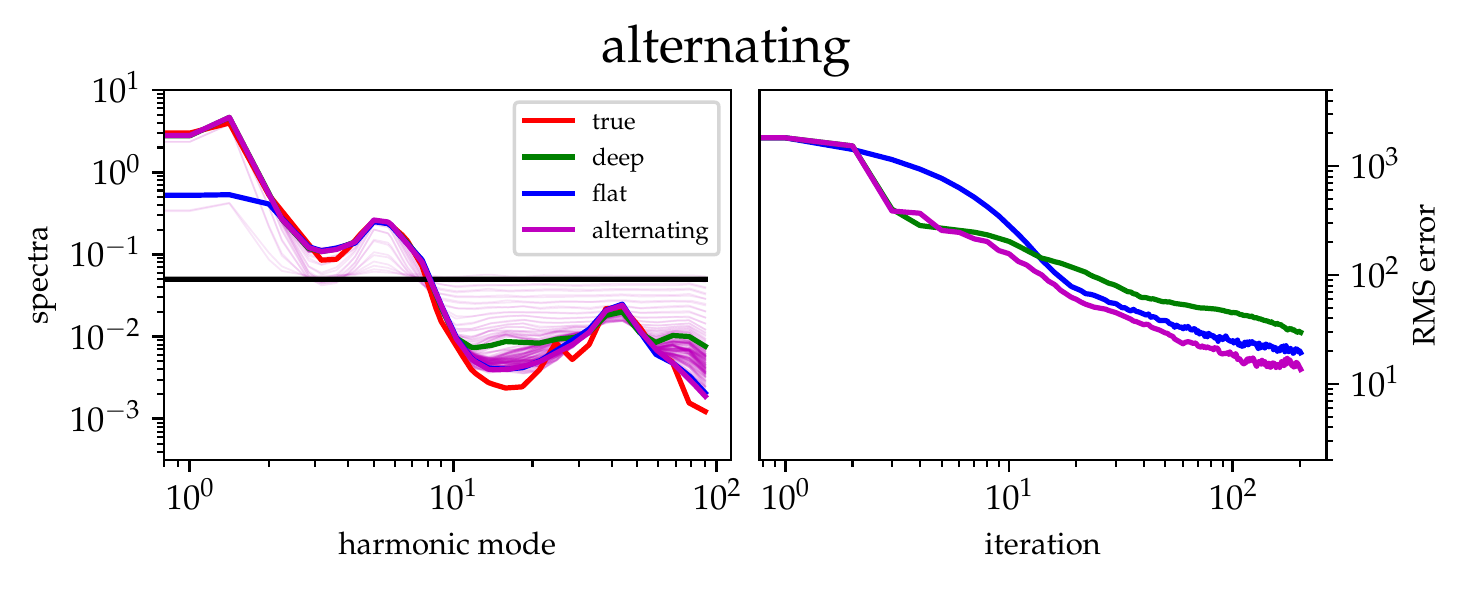}
\caption{The results for alternating between the two parametrizations during the inference. Left is the progression of the power spectra and right the RMS error to the Wiener filter.}
\label{fig:alternating}
\end{figure}

Alternating between both methods might allow to overcome the limitations of either one, leading to overall faster convergence and more reasonable spectra. We will restrict our investigation to the scattered data case, where the limitations of both methods were most apparent. The results can be seen in Fig. \ref{fig:alternating}. In terms of convergence, alternating the procedures allows to exceed the performance of both methods on their own. The alternating method follows the initial drop of the deep hierarchy model and for the rest it behaves as the flat model, providing the overall lowest error. This can also be seen in the progression of the spectra. The large scales behave like the deep model, rapidly jumping to the correct value, but also match the flat model behavior on the smallest scales. 

Overall, the variational approximation in the standardized coordinates demonstrates its numerical superiority in the case of a weak likelihood, or more general, in directions the likelihood is relatively uninformative. The overall convergence is limited by the best informed directions, but the other directions converge fast nevertheless. The deep hierarchical parametrization shows the opposite behavior and a MAP estimate for well constrained parameters allows us to alternate between the parametrization, allowing for overall faster convergence.

\section{Conclusion}
\label{sec:conclusion}
We showed how the multivariate distributional transform can be used to transform a general Bayesian model over continuous quantities into a standardized coordinate system in which the prior becomes a standard Gaussian distribution. We discussed the behavior of popular approximations in these new coordinates. A maximum posterior approximation in this space will be oriented towards median prior configurations in the original space, instead of the maximum. For a strong likelihood, the resulting distributions will be transformed versions of each other. A variational Gaussian approximation in the standardized coordinates will be excellent in cases of a weak likelihood, as the shape of the posterior will only be slightly modified compared to the prior, and therefore a Gaussian approximation can capture the true posterior well. This is also the case if the likelihood, containing the standardizing transformation, is still close to a Gaussian. 

The simple structure of the model in the standardized coordinates allowed us to investigate the numerical behavior of optimization schemes in terms of the conditioning of the curvature. Its smallest eigenvalue is larger or equal one, and the overall conditioning mainly depends on the largest eigenvalue of the likelihood. This eigenvalue is large if the data contains large amounts of information on certain parameter directions, and small otherwise. This makes the inference of the approximation parameters easier for small amounts, or uninformative data. This makes a Gaussian approximation in the standardized coordinates, as proposed for ADVI \citep{ADVI} a fast and accurate inference procedure especially in cases of sparse data.

We explored numerically the inference of a Gaussian process with unknown correlation structure, which was described by a smooth  log-Gaussian process, this time with known, smoothness enforcing kernel. The linearity of the standardizing transformation allowed us to perform the identical approximation in the original, as well as the standardized coordinates. The approximation was a variational Gaussian with full covariance for the Gaussian process, and a point estimate for the power spectrum.
We found, as expected, that the flat hierarchical model was superior for less informed parameters, whereas the deep hierarchy outperformed for well-constrained parameters. 
Both scenarios occur also within the same inference problem, and the convergence speed of either approximation is bottle-necked by its sub-optimal directions.
We solved this problem by alternating the optimization between the two coordinate systems, which harnessed the strength of both and lead to an overall faster convergence. We proposed that parameters well constrained by the data should be approximated by delta distributions, which allows their inference in either coordinates. This should improve the inference speed of large, hierarchical Bayesian models.

\section{Acknowledgment}
We acknowledge Philipp Arras, Philipp Frank, Maksim Greiner, Sebastian Hutschenreuter, Reimar Leike and Martin Reinecke for fruitful discussions and remarks on the manuscript.
\appendix

\section{A simple transformation example}
\label{app:simple}
We perform this transformation explicitly to a one dimensional example with a hierarchical Bayesian model with two parameters. 
We consider some likelihood $\mathcal{P}(d\vert \alpha)$ that depends on a parameter $\alpha$. The prior distribution on this is a Gaussian with the standard deviation $\sigma$ and a known mean $\mu$ as hyper-parameters. The hyper-prior on this will be an exponential distribution depending on a known constant $\lambda$. A similar example is discussed in \citet{HMCHierarchy}.
\begin{align}
\mathcal{P}(\alpha\vert \sigma) =& \: \mathcal{G}(\alpha-\mu, \sigma^2) \qquad\qquad \alpha, \mu \in \mathbb{R}\\
\mathcal{P}(\sigma) =& \: \lambda e^{-\lambda \sigma} \qquad \qquad \qquad  \quad\sigma, \lambda \in \mathbb{R}^+
\end{align}
To calculate the transformations onto white priors we require the CDF of a white Gaussian, which is stated in Eq. \ref{eq:CDFGaussian}, as well as the inverse CDF's of the prior distributions:
\begin{align}
\mathcal{F}^{-1}_{\mathcal{P}(\alpha\vert \sigma)} (u)=& \: \mu + \sqrt{2}\sigma \: \mathrm{erf}^{-1}\left( 2 u - 1\right) \\
\mathcal{F}^{-1}_{\mathcal{P}(\sigma)} = & \:- \frac{1}{\lambda} \mathrm{ln}(1-u)
\end{align}
We can then express our encoding of the prior structure in the likelihood as
\begin{align}
\alpha =& \: \mathcal{C}_{\mathcal{P}(\alpha\vert \sigma)}(\xi_1) = \mathcal{F}^{-1}_{\mathcal{P}(\alpha\vert \sigma)} \circ \mathcal{F}_{\mathcal{G}(\xi,\mathbb{1})}(\xi_1)\\
= & \: \mu + \sqrt{2}\sigma \: \mathrm{erf}^{-1}\left( 2 \left(\frac{1}{2} + \frac{1}{2} \mathrm{erf}\left( \frac{\xi_1}{\sqrt{2}}\right)\right) - 1\right) \\
=& \: \mu + \sigma \xi_1 \text{.}
 \label{eq:1dgauss}
\end{align}
This is just a simple re-scaling and shifting of the white Gaussian $\xi$ parameter within the likelihood. This is exactly the example given for the reparametrization trick \citep{AEVB} and the proposed transformation for hierarchical HMC \citep{HMCHierarchy}.

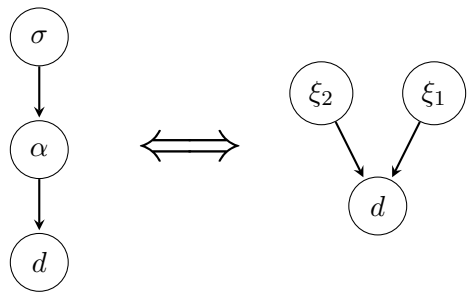
\begin{figure}[h]
	\centering
	\begin{tikzpicture}
	[c/.style={circle,minimum size=2em,text centered,thin},
	r/.style={rectangle,minimum size=2em,text centered,thin},
	v/.style={->,shorten >=1pt,>=stealth,thick}, 
	arrow/.style={-latex, shorten >=1ex, shorten <=1ex, bend angle=45}]
	\node(b)at(-3.25,5)[c,draw]{$\sigma$};
	\node(c)at(-3.25,3.5)[c,draw]{$\alpha$};
	\node(d)at(-3.25,2)[c, draw]{$d$};
	\node(e)at(2,4.25)[c,draw]{$\xi_1$};
	\node(f)at(0.5,4.25)[c,draw]{$\xi_2$};
	\node(g)at(1.25,2.75)[c, draw]{$d$};
	\node(h)at(-1.25,3.5)[scale=2]{$\Longleftrightarrow$};
	\draw[v](b.south)--(c);
	\draw[v](c.south)--(d);
	\draw[v](e)--(g);
	\draw[v](f)--(g);	    	  
	\end{tikzpicture}
	\flushleft
	\caption{The graphical structure of the original model with a deep hierarchy and the flattened structure of the transformed model.}
	\label{fig:diagram} 
\end{figure}

Analogously we perform the transformation for the second parameter $\sigma$:
\begin{align}
\sigma =& \: \mathcal{C}_{\mathcal{P}(\sigma)}(\xi_2)\\ 
=& \:-\frac{1}{\lambda}  \mathrm{ln}\left( \frac{1}{2}-\frac{1}{2}\mathrm{erf}\left(\frac{\xi_2}{\sqrt{2}}\right)\right)
\end{align}
With this we can substitute $\sigma$ in the likelihood to obtain full white prior distributions. We can relate back to the original parameter of the likelihood via
\begin{align}
\alpha =\mu -\frac{1}{\lambda}  \mathrm{ln}\left( \frac{1}{2}-\frac{1}{2}\mathrm{erf}\left(\frac{\xi_2}{\sqrt{2}}\right)\right) \xi_1
\end{align}
and the posterior can be written as
\begin{align}
\mathcal{P}(\xi_1,\xi_2\vert d) = \frac{\mathcal{P}(d\vert \xi_1, \xi_2)\mathcal{P}(\xi_1)\mathcal{P}(\xi_2)}{\mathcal{P}(d)}
\end{align}
instead of 
\begin{align}
\mathcal{P}(\alpha,\sigma \vert d) = \frac{\mathcal{P}(d\vert \alpha)\mathcal{P}(\alpha\vert \sigma)\mathcal{P}(\sigma)}{\mathcal{P}(d)} \text{.}
\end{align}
A graphical representation of the of the conditional dependence of the variables within the two models, the initial and the one re-parametrized to have white Gaussian random variables, is depicted in Fig. \ref{fig:diagram}. With this transformation we flattened down the hierarchy of the original, deep Bayesian model. Both models contain the identical information.  This is now encoded in the nonlinear structure of the parameters within the likelihood. One could continue adding higher levels of prior hierarchies to the problem, but flattening them as well is straight forward, as long as the CDF's are available. As it is pointed out in \citet{HMCHierarchy}, the parameters are now independent, conditioned on the data.

\section{Transforming multivariate Gaussians}
\label{app:multivariate}
Multivariate Gaussian distributions and their generalizations to infinite dimensions, Gaussian processes \citep{krige1951statistical,GPML} are an important class of distributions, especially to express prior knowledge. They take the following form:
\begin{align}
\mathcal{P}(s) = \mathcal{G}(s,S) = \frac{1}{\vert 2 \pi S\vert^{\frac{1}{2}}} e^{-\frac{1}{2}s^\dagger S^{-1} s}
\end{align}
Here we do not have to perform the transformation explicitly, as we can find the white, standard Gaussian parametrization through a set of linear transformations. 
First, we have to express the correlation structure in terms of its eigenbasis.
\begin{align}
S = \mathbb{F^\dagger} \widetilde{S} \mathbb{F}
\end{align}
Here the unitary transformation $\mathbb{F}$ is built from the normalized eigenvectors and the diagonal matrix $\widetilde{S}$ consists of the corresponding eigenvalues. The inverse in this basis can be calculated easily by inverting each individual eigenvalue and it reads
\begin{align}
S^{-1} = \mathbb{F^\dagger} \widetilde{S}^{-1} \mathbb{F} \text{.}
\end{align}
With this we can rewrite the exponent of the Gaussian distribution as 
\begin{align}
\frac{1}{2} s^\dagger S^{-1} S =\frac{1}{2} s^\dagger \mathbb{F}^\dagger \widetilde{S}^{-1} \mathbb{F} s \:\equiv \:\frac{1}{2} \widetilde{s}^\dagger \widetilde{S}^{-1} \widetilde{s} \text{.}
\end{align} 
The eigenvalues of $S$ encode the variance of the process along their corresponding eigendirections and it is the squared standard deviation. We can split up this diagonal covariance into two amplitude matrices by taking the matrix square root:
\begin{align}
\frac{1}{2} \widetilde{s}^\dagger \sqrt{\widetilde{S}^{-1}}^\dagger\sqrt{\widetilde{S}^{-1}}  \widetilde{s} = \frac{1}{2} \widetilde{s}^\dagger \sqrt{\widetilde{S}^{-1}}^\dagger \mathbb{1}\sqrt{\widetilde{S}^{-1}}  \widetilde{s} 
\end{align}
Finally, we can introduce the new variable $\xi =\sqrt{\widetilde{S}^{-1}}  \widetilde{s}$ and the exponential of this Gaussian distribution becomes
\begin{align}
\frac{1}{2}s^\dagger S^{-1} s = \frac{1}{2} \xi^\dagger \mathbb{1} \xi
\end{align}
with the full transformation 
\begin{align}
s = \mathbb{F} \sqrt{\widetilde{S}} \xi \equiv A \xi \text{ .}
\end{align}
This is therefore analogous to the one dimensional case as given in Eq. \ref{eq:1dgauss}. We weight white, random excitations $\xi$ in the eigenbasis of the correlation structure with the corresponding amplitudes contained within $A$ in terms of the standard deviation and transform it back to the space we are interested in.
The original correlation structure is expressed in terms of it as $S = A A^\dagger$.

\bibliography{citations}

\end{document}